\documentclass{article}
\usepackage{ijcai23}

\usepackage{times}
\usepackage{soul}
\usepackage{url}
\usepackage[hidelinks]{hyperref}
\usepackage[utf8]{inputenc}
\usepackage[small]{caption}
\usepackage{graphicx}
\usepackage{amsmath}
\usepackage{amsthm}
\usepackage{booktabs}
\usepackage{algorithm}
\usepackage{algorithmic}
\usepackage[switch]{lineno}

\usepackage{amssymb}
\usepackage{helvet}
\usepackage{courier}
\usepackage{epsfig}
\usepackage{mathrsfs}
\usepackage{multirow}
\usepackage{color}
\usepackage{xcolor}
\usepackage{colortbl}
\usepackage{tabularx}
\usepackage{gensymb}
\usepackage{arydshln}

\definecolor{Gray}{gray}{0.95}
\usepackage{colortbl}
\newcolumntype{a}{>{\columncolor{Gray}}c}

\newcommand{\ljx}[1]{\textcolor{black}{#1}}
\newcommand{\seasons}[1]{\textcolor{black}{#1}}

\newcommand{\jinxiang}[1]{\textcolor{black}{#1}}

\title{Clustered-patch Element Connection for Few-shot Learning}

\author{
\quad\quad\  Jinxiang Lai$^1$\and
Siqian Yang$^1$\and
Junhong Zhou$^2$\and
Wenlong Wu$^1$\and
Xiaochen Chen$^1$\and
\newline
Jun Liu$^1$\and
Bin-Bin Gao$^{*1}$\and
Chengjie Wang\thanks{Corresponding Author} $^{1, 3}$
\affiliations
$^1$Tencent Youtu Lab, China\\
$^2$Southern University of Science and Technology, China\\
$^3$Shanghai Jiao Tong University, China
\emails
layjins1994@gmail.com, \{seasonsyang, ezrealwu, husonchen\}@tencent.com, 12011801@mail.sustech.edu.cn, \{junsenselee, csgaobb\}@gmail.com, jasoncjwang@tencent.com
}

\begin{document}

\maketitle

\begin{abstract}
Weak feature representation problem has influenced the performance of few-shot classification task for a long time.
To alleviate this problem, recent researchers build connections between support and query instances through embedding patch features to generate discriminative representations.
\seasons{However, we observe that there exists semantic mismatches (foreground/ background) among these local patches, because the location and size of the target object are not fixed.}
What is worse, these mismatches result in unreliable similarity confidences, and complex dense connection exacerbates the problem.
According to this, we propose a novel Clustered-patch Element Connection (CEC) layer to correct the mismatch problem.
The CEC layer leverages Patch Cluster and Element Connection operations to collect and establish reliable connections with high similarity patch features, respectively.
\jinxiang{Moreover, we propose a CECNet, including CEC layer based attention module and distance metric.}
The former is utilized to generate a more discriminative representation benefiting from the global clustered-patch features, and the latter is introduced to reliably measure the similarity between pair-features.
Extensive experiments demonstrate that our CECNet outperforms the state-of-the-art methods on classification benchmark.
\jinxiang{Furthermore, our CEC approach can be extended into few-shot segmentation and detection tasks, which achieves competitive performances.}
\end{abstract}

%\vspace{-0.4cm}
\section{Introduction}
\label{sec:Introduction}
%\vspace{-0.1cm}
%Contrast to general deep learning task \cite{krizhevsky2012imagenet,simonyan2015very}, \textit{Few-Shot Learning} (FSL) aims to learn a transferable classifier with amount seen images (base class) and few labeled unseen images (novel class).
%Due to the lack of effective features from unseen class, a robust feature embedding model is indispensable.
%Recent researchers\cite{hou2019cross,tian2020rethinking,rizve2021exploring,xu2021learning,ye2020feat} manage to design an embedding network for generating more discriminative features.
In contrast to general deep learning task \cite{krizhevsky2012imagenet}, \textit{Few-Shot Learning} (FSL) aims to learn a transferable classifier with amount seen images (base class) and few labeled unseen images (novel class).
Due to the lack of effective features from unseen classes, a robust feature embedding model is indispensable.
Recent researchers\cite{hou2019cross,rizve2021exploring,xu2021learning} manage to design an embedding network for generating more discriminative features.

\begin{figure}[!t]
\centering
\includegraphics[width=0.99\linewidth]{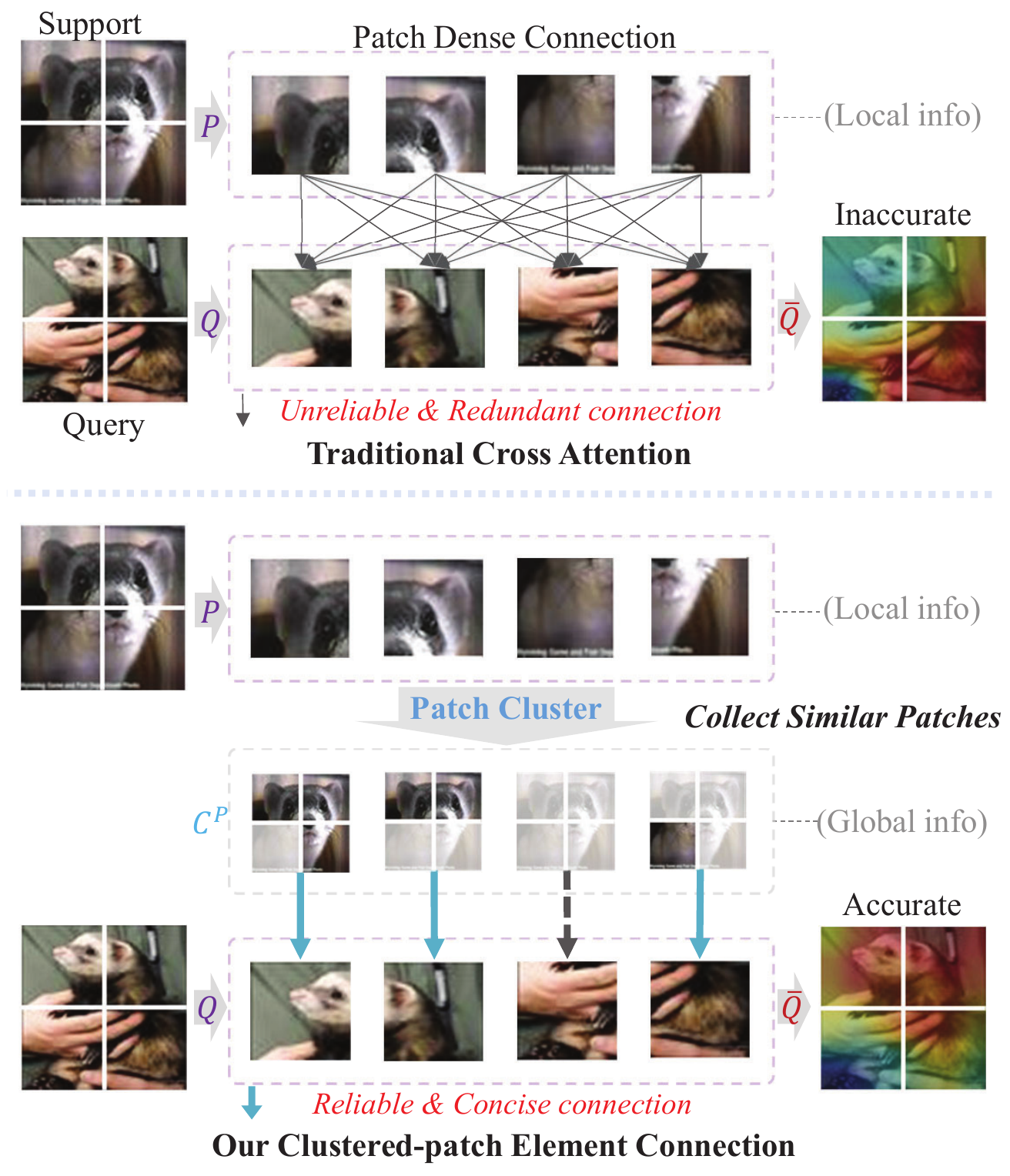}
%\vspace{-2mm}
\caption{Comparison between traditional Cross Attention and our Clustered-patch Element Connection.
The proposed Clustered-patch Element Connection, which utilizes the global info $C^p$ integrated from support feature $P$ to perform element connection with query $Q$ leading to a confident and clear connection, is able to generate a more clear and precise relation map than Cross Attention.
The detailed Patch Cluster operation is illustrated in Fig.\ref{fig:patch}. The visualization comparisons are referred to Fig.\ref{fig:visual}(a).}
\label{fig:motivation}
%\vspace{-4mm}
\end{figure}

\begin{figure}[!t]
\centering
\includegraphics[width=0.99\linewidth]{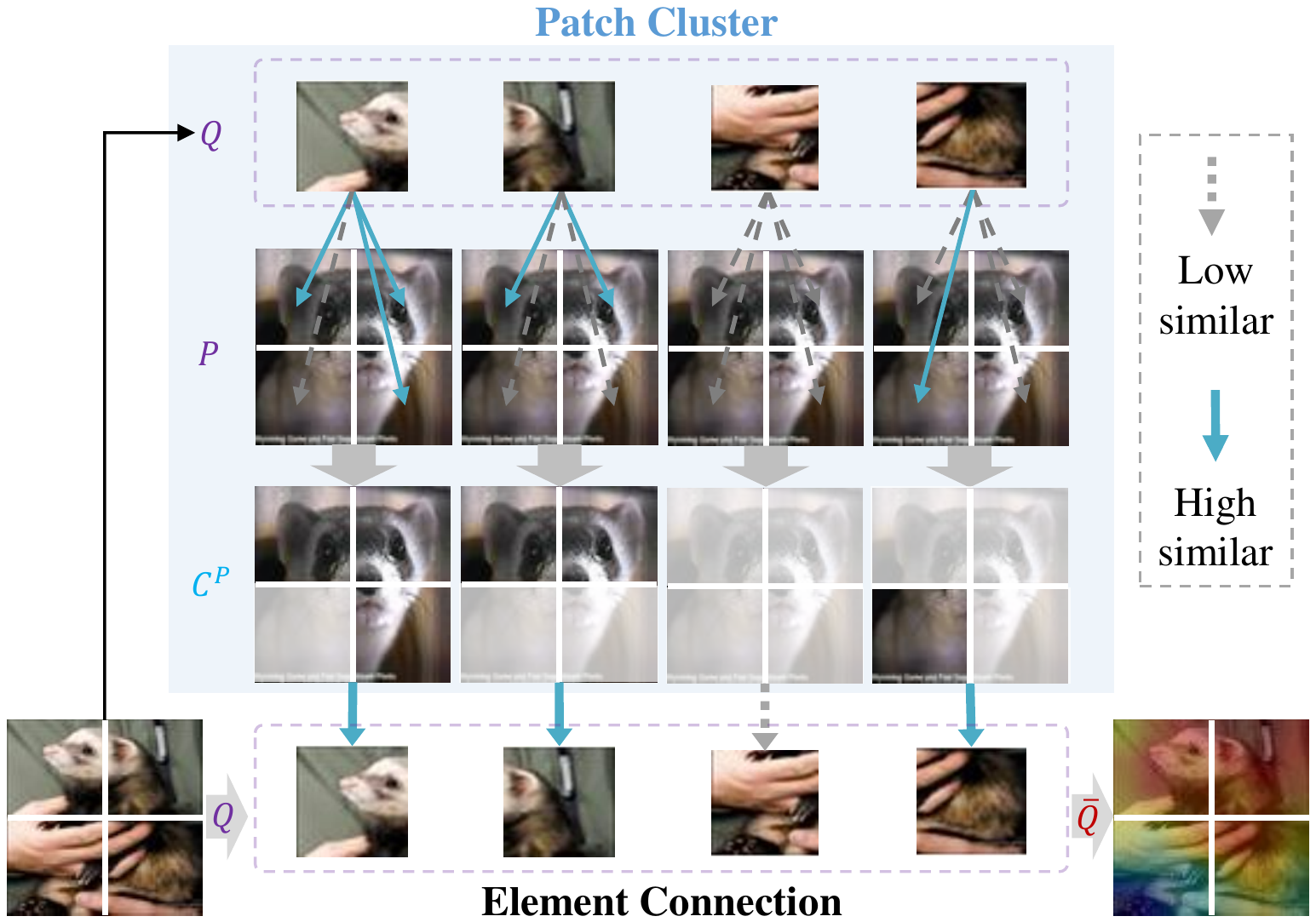}
%\vspace{-1mm}
\caption{Patch Cluster and Element Connection.}
\label{fig:patch}
%\vspace{-0.4cm}
\end{figure}

Specifically, cross attention based methods \cite{hou2019cross,xu2021learning,xu2021Rectifying} focus on reducing the background noise and highlighting the target region to generate more discriminative representations.
The core idea of these methods is to divide extracted features into patches and connect all local patch features.
However, \ljx{as shown in Fig.\ref{fig:motivation},} we observe that the target object may be located randomly with different scales among the query images. Hence, these methods suffer \seasons{two main problems: inconsistent semantic} in feature space, and \seasons{unreliable and redundant} connections.
To tackle these problems, we propose a Clustered-patch Element Connection (CEC) layer which consists of Patch Cluster and Element Connection operations.
In detail, given inputs features ${P}$ (support) and ${Q}$ (query), CEC layer firstly obtains the global clustered-patch ${C^p}$ features by Patch Cluster operation \ljx{as illustrated in Fig.\ref{fig:patch}}, then performs Element Connection on ${Q}$ by using ${C^p}$, and finally produces a more discriminative representation ${\bar{Q}}$.
Patch Cluster aims to collect the objects in source feature ${P}$ that are similar to the reference patch in ${Q}$, which adaptively alignments the ${P}$ into ${C^P}$ to obtain a consistent semantic feature for each patch of ${Q}$. Then, with the global clustered-patch features, CEC layer generates more reliable and concise connections than cross attention.

According to CEC layer, we find the key of generating accurate relation map is to obtain appropriate clustered-patch features. \ljx{In this paper}, four solutions are introduced to perform Patch Cluster, including MatMul, Cosine, GCN and Transformer. Different from the naive MatMul and Cosine modes, we propose the meta-GCN and Transformer based Patch Cluster operations to obtain a more robust clustered-patch by implementing additional feature refinement.
The insight of meta-GCN is constructing a dynamic correlation-based adjacent for each current input pair-features, other than the static GCN \cite{kipf2017semi} using a fixed adjacent.
Besides, the transformer structure obtains global information via modeling a spatio-temporal correlation among instances, which generates a more accurate relation map.

Along with the description of CEC mechanism, we propose three CEC-based modules:
(I) The Clustered-patch Element Connection Module (CECM) distinguishes the background and the object for each image pair (support and query) at the feature level adaptively, which gives a more precise highlights at the regions of target object;
(II) The Self-CECM enhances the semantic feature of target object in a self-attention manner to make the representation more robust;
(III) The Clustered-patch Element Connection Distance (CECD) is a CEC-based distance metric which measures the similarity between pair-features via the obtained reliable relation map.

For few-shot classification task, we introduce a novel Clustered-patch Element Connection Network (CECNet) as illustrated in Fig.\ref{fig:CECN}, which learns a generalize-well embedding benefiting from auxiliary tasks, generates a discriminative representation via CECM and Self-CECM, and measures a reliable similarity map via CECD.
\jinxiang{Furthermore, we derive a novel CEC-based embedding module named CEC Embedding (CECE), which can be applied into few-shot semantic segmentation (FSSS) and few-shot object detection (FSOD) tasks.
We simply stack the proposed CECE after the backbone network of the existing FSSS and FSOD methods, which achieves consistent improvements around $1\%-3\%$.}

To summarize, our main contributions are:

$\bullet$ We propose a Clustered-patch Element Connection (CEC) layer to strengthen the target regions of query features by element-wisely connecting them with the global clustered-patch features. Four different CEC modes are introduced, including MatMul, Cosine, GCN and Transformer.

$\bullet$ We derive three CEC-based modules: CECM and Self-CECM modules are utilized to produce more discriminative representations, and CECD is able to measure a reliable similarity map.

$\bullet$ With CEC-based modules and auxiliary tasks, a novel CECNet model is designed for few-shot classification.
CECNet improves state-of-the-arts on few-shot classification benchmark, and the experiments demonstrate that our method is effective in FSL.

$\bullet$ \jinxiang{Furthermore, our CECE (i.e. CEC-based embedding module) can be extended into few-shot segmentation and detection tasks, which achieves performance improvements around $1\%-3\%$ on the corresponding benchmarks.}

%\vspace{-0.1cm}
\section{Related Work}
\paragraph{Few-Shot Learning}
\jinxiang{The FSL algorithms aim to recognize novel categories with few labeled images, and a category-disjoint base set with abundant images is provided for pre-training.
The classic FSL tasks include few-shot classification \cite{finn2017model,vinyals2016matching,snell2017prototypical,hou2019cross,tian2020rethinking}, semantic segmentation \cite{zhang2020sg,siam2019amp,Malik2021repri} and object detection \cite{kang2019few,wang2020frustratingly,qiao2021defrcn}.
More introductions are presented in APPENDIX.
In a word, the existing FSL methods lack a uniform function to control the connections among the patches between support and query instances semantically.}

%\seasonsyang{The representative FSL approaches can be divied into four aspects: parameter-generating based \cite{munkhdalai2017meta,gidaris2019generating}, optimization-based \cite{nichol2018first,finn2017model,marcin2018learn,ravi2016optimization}, metric-learning based \cite{vinyals2016matching,snell2017prototypical,xu2021learning,hou2019cross}, and embedding-based methods \cite{tian2020rethinking,rizve2021exploring,zhengyu2021pareto}. }
%\jinxiang{Due to the limitation of the space, more introductions of these FSL algorithms are presented in APPENDIX.
%In a word, the existing few-shot learning methods lack of an uniform function to control the connections among the patches between support and query instances semantically.}

% \jinxiang{The representative FSL classification approaches include parameter-generating based \cite{munkhdalai2017meta,gidaris2019generating}, optimization-based \cite{nichol2018first,finn2017model,marcin2018learn,ravi2016optimization}, metric-learning based \cite{vinyals2016matching,snell2017prototypical,xu2021learning,hou2019cross}, and embedding-based methods \cite{tian2020rethinking,rizve2021exploring,zhengyu2021pareto}.
% More introductions of these FSL algorithms are presented in APPENDIX.
% In a word, the existing few-shot classification methods lack of an uniform function to control the connections among the patches between support and query instances.
% }

\paragraph{Other Related Works}
\jinxiang{are introduced in APPENDIX, such as \textbf{Auxiliary Task for FSL} \cite{hou2019cross,rizve2021exploring,lai2022rethinking,lai2022tsf}, \textbf{Graph Convolutional Network (GCN)} \cite{bruna2013spectral}, and \textbf{Transformer} \cite{vaswani2017attention}.}

%\vspace{-0.1cm}
\section{Problem \seasons{Definition}}
%\vspace{-0.1cm}
\subsection{Few-Shot Classification}
\label{sec:ProblemDef}
\seasons{A classic few-shot classification problem is specified as a ${N}$-way ${K}$-shot task, which means solving a $N$-class classification problem with only $K$ labeled instances provided per class.
In the recent investigations\cite{hou2019cross,snell2017prototypical}, the source dataset is divided into three category-disjoint parts: training set ${X_{train}}$, validation set ${X_{val}}$ and test set ${X_{test}}$.
Moreover, the episodic training mechanism is widely adopted.
An episode consists of two sets (randomly sampling in $N$ categories): support and query. Let $\mathcal{S}=\{\left(x^s_i, y^s_i\right)\}_{i=1}^{n_s}$ ($n_s=N\times K$) denote the support set, and $\mathcal{Q}=\{\left(x^q_i, y^q_i\right)\}_{i=1}^{n_q}$ denote the query set.
Note that $n_{s}$ and $n_{q}$ are the size of corresponding sets.
Especially, $\mathcal{S}=\{\mathcal{S}^1, \mathcal{S}^2, ..., \mathcal{S}^k\}$, where $\mathcal{S}^k$ denotes the support set of the $k^{th}$ category in $\mathcal{S}$.}
\ljx{\seasons{Specifically}, let ($P$, $Q$) $\in \mathbb{R}^{hw \times c}$ denote the support and query features, which are extracted from support subset and query instance ($\mathcal{S}^k$, $x^q$). Note that $c,$ $h$, $w$ are channel number, height, width of features, respectively.}

\subsection{Cross Attention}
The traditional Cross Attention \cite{hou2019cross} proves that highlighting target regions \seasons{could} generate more discriminative representations, leading to accuracy improvements for FSL.
The key is \seasons{to generate} an \seasons{fine-grained} relation map ${R^Q}\in \mathbb{R}^{hw}$ \seasons{to represent} the target regions \seasons{in} $Q\in \mathbb{R}^{hw \times c}$.
\seasons{Then, a spatial-wise feature attention can be obtained} through $R^Q \odot Q$, where ${\odot}$ is the Element-wise Product.
The traditional Cross Attention produces relation map ${R^Q}$ for ${Q}$ by:
\begin{small}
\begin{equation}
%\vspace{-0.1cm}
{R^Q} = h\left(\frac{P}{||P||_2} \left(\frac{Q}{||Q||_2}\right)^T\right),
\label{eq:cam}
%%\vspace{-0.1cm}
\end{equation}
\end{small}where ${h}$ is a CNN-based layer to refine the correlation matrix ${\frac{P}{||P||_2} (\frac{Q}{||Q||_2})^T\in \mathbb{R}^{{hw}\times {hw}}}$.
According to Eq.\ref{eq:cam}, the Cross Attention produces relation map by \seasons{\emph{Local-to-Local}} fully connection among local feature patches of $P$ and $Q$.
In detail, $({P_i},{Q_j}) \in \mathbb{R}^{1 \times c}$ represent \seasons{a pair of} support feature patch and query feature patch among $({P},{Q}) \in \mathbb{R}^{hw \times c}$ .
As shown in Fig.\ref{fig:motivation}, the target object may be located \seasons{unregularly} among the query images at different scale, which results in inconsistent semantic in feature space, i.e feature patches ${P_i}$ and ${Q_j}$ may be semantically inconsistent.
This \emph{semantically inconsistent problem} causes low confident correlation between patches, and the complex Local-to-Local fully connection further accumulates this inaccurate bias, which affect the quality of the generated relation map.

\seasons{To establish concise and clear connections among global and local features}, we propose a Clustered-patch Element Connection layer (CEC), \seasons{which consists of two key operations: \emph{Patch Cluster} and \emph{Element Connection}.}

%\vspace{-0.1cm}
\section{Clustered-patch Element Connection}\label{Clustered-patch_Element_Connection}
%\vspace{-0.1cm}
\subsection{Patch Cluster}
\ljx{As illustrated in Fig.\ref{fig:patch}}, Patch Cluster \seasons{operation} \ljx{obtains} a set \ljx{$C^p$}, \seasons{named Clustered-patch}, \ljx{via collecting} those objects in support feature set ${P}$, which are similar to the reference patch in ${Q}$.
We define a generic Patch Cluster operation $f_{PC}$ as:
\begin{equation}
{C^p} = f_{PC}(Q,P)=\phi\left(g\left(Q,P\right)P\right).
\label{eq:pc}
\end{equation}
Here ${P}$ is the input source feature, ${Q}$ is the input reference feature, and ${C^p} \in \mathbb{R}^{hw \times c}$ is the output Clustered-patch. A pairwise function ${g}$ computes an affinity matrix representing relationship between ${Q}$ and ${P}$. The clustered patches can be refined by function ${\phi}$.
\seasons{
In detail, we divide the source image into $w\times{h}$ patches.
Here, $w$ and $h$ are the same as the size of the features in $P$, which is convenient for element connection operation.
A Clustered-patch ${C^p} \in \mathbb{R}^{hw \times c}$ collects $w\times{h}$ clusters.
Each cluster collects the patch-features in ${P}=[P_1,P_2,...,P_{hw}] \in \mathbb{R}^{hw \times c}$ that are similar to the corresponding patch-feature in the reference patch-features in $Q$.}
Therefore, ${C^p}$ is semantically similar to ${Q}$.

\seasons{To implement the Patch Cluster operation, we give four solutions including MatMul, Cosine, GCN and Transformer.}

\paragraph{MatMul}
A simplest way to obtain the clustered patches is \seasons{treating} MatMul operation as the pairwise function ${g}$ (in Eq.(\ref{eq:pc})) and not implementing any further embedding refinement. Formally,
\begin{small}
\begin{equation}
{C^p} = \sigma\left({Q}{P^T}\right)P,
\label{eq:matmul}
\end{equation}
\end{small}where ${\sigma}$ is softmax function.

\paragraph{Cosine}
A simple extension of the MatMul version is to compute cosine similarity in feature space. Formally,
\begin{small}
\begin{equation}
{C^p} = \sigma\left(\frac{Q}{||Q||_2} \left(\frac{P}{||P||_2}\right)^T\right)P.
\label{eq:cosine}
\end{equation}
\end{small}

\paragraph{GCN}
GCN \cite{kipf2017semi} updates the input features ${P}$ via utilizing a pre-defined adjacent matrix ${A}\in\mathbb{R}^{{hw}\times{hw}}$ and a learnable weight matrix ${W}\in\mathbb{R}^{{c}\times{c}}$.
Formally, the updated features ${G^p}\in\mathbb{R}^{{hw}\times{c}}$ can be expressed as: ${G^p} = \delta({A}{P}{W})$, where $\delta(\cdot)$ is the nonlinear activation function ($Sigmoid(\cdot)$ or $ReLU(\cdot)$).
However, the adjacent matrix ${A}$ used in GCN is fixed for all inputs after training, which is not able to recognize the new categories in few-shot task. Comparing Eq.(\ref{eq:pc}) and the definition of GCN, we observe that the affinity matrix $g\left(Q,P\right)$ can be considered as the adjacent matrix ${A}$,  \seasons{because they all try to describe the relationship between features $P$ and $Q$.}
\seasons{Hence, we derive a meta-GCN through replacing the static adjacent matrix with the dynamic affinity matrix.}
Formally, the meta-GCN based Patch Cluster operation is derived as follows:
\begin{small}
\begin{equation}
{C^p} = \delta\left[\sigma\left(\frac{Q}{||Q||_2} \left(\frac{P}{||P||_2}\right)^T\right)PW\right].
\label{eq:metagcn}
\end{equation}
\end{small}

\paragraph{Transformer}
\seasons{The Transformer\cite{vaswani2017attention} based Patch Cluster operation is defined as follows:}
\begin{small}
\begin{equation}
C^p=FFN\{\sigma[({W_q}Q)({W_k}P^T)]{W_v}P\},
\label{eq:cecm_t}
\end{equation}
\end{small}where, $FFN$ is the Feed-Forward Network in transformer, ${W_{q},W_{k},W_{v}}$ are learnable weights (e.g. convolution layers).

\subsection{Element Connection}
\seasons{According to the global semantic features ${C^p}$ obtained from Patch Cluster operation, } element Connection operation generates the relation map ${R^Q}$ for ${Q}$ by simply computing the patch-wise cosine similarity between ${Q}$ and ${C^p}$.
Finally, we obtain a rectified discriminative representation by the Element Connection operation $f_{EC}$:
\begin{small}
\begin{equation}
\begin{aligned}
{\bar{Q}}&=f_{EC}(Q,C^p)= \left(\sigma\left({R^Q}\right) + 1\right)\odot Q,\\
&where,\quad {R^Q}= \left(\frac{Q}{||Q||_2}\otimes \frac{C^p}{||C^p||_2}\right)\in \mathbb{R}^{hw},
\label{eq:ec_Q}
\end{aligned}
\end{equation}
\end{small}where, $\otimes$ is Patch-wise Dot Product, ${\odot}$ is Element-wise Product. The ${n^{th}}$ position of ${R^Q}$ is $R^Q_n=\frac{Q_{n}}{||Q_{n}||_2}\cdot\frac{C^p_{n}}{||C^p_{n}||_2}$, where $\cdot$ is Dot Product.
The visualizations of the CEC-based relation map ${R^Q}$ are shown at the last column in Fig.\ref{fig:visual}(b).
\ljx{
Overall, the Clustered-patch Element Connection (CEC) layer is able to highlight the regions of $Q$ that are semantically similar to $P$. Formally, CEC layer $f_{CEC}$ is expressed as:
\begin{equation}
\begin{aligned}
{\bar{Q}} = f_{CEC}(Q,P)=f_{EC}\left(Q,f_{PC}(Q,P)\right).
\label{eq:cec}
\end{aligned}
\end{equation}
}

%\vspace{-0.4cm}
\subsection{Discussion}
\seasons{Compared with traditional Cross Attention, the key point of our Clustered-patch Element Connection is to perform the Global-to-Local element connection between the Clustered-patch $C^p$ (global) and query $Q$ (local).
It is able to generate a more clear and precise relation map, as shown in Fig. \ref{fig:visual}(a) visualizations.
As demonstrated in Tab.~\ref{table:ablation_cecm}, our CEC-based approach achieves 4\% accuracy improvement than the traditional Cross Attention based CAN \cite{hou2019cross}.}

Generally, the advantages of our Clustered-patch Element Connection are: (I) The relation map generated by Element Connection is more confident than Cross Attention, because the global Clustered-patch feature ${C^p}$ is more stable and representative than the local feature ${P}$. (II) Element Connection (1-to-1 patch-connection) has more clear connection relationship than Cross Attention (1-to-${hw}$ patch-connection).

Moreover, the respective advantages of different solutions for realizing Patch Cluster are: (I) These four solutions can be divided into two groups: fixed (i.e. MatMul and Cosine) and learnable (i.e. GCN and Transformer) solutions. The fixed solutions can be used to perform patch clustering without additional learnable parameters, while the learnable solutions are data-driven to refine the affinity matrix or clustered-patch.
(II) According to experimental results in Tab. \ref{table:ablation_cecm}, the learnable solutions are better than the fixed ones when they are applied as a embedding layer for feature enhancing (\ljx{i.e. CECM defined in Eq.~\ref{eq:cecm}}), which indicates that the learnable solutions can generate better embedding features.
\seasons{In contrast,} according to Tab. \ref{table:ablation_cecd}, the fixed solutions are better than the learnable ones when they are applied as the distance metric for measuring similarity (\ljx{i.e. CECD defined in Eq.~\ref{eq:cecd}}), which indicates fixed solutions can obtain more reliable similarity scores.

\begin{figure*}[!t]
\centering
\includegraphics[width=0.9\linewidth]{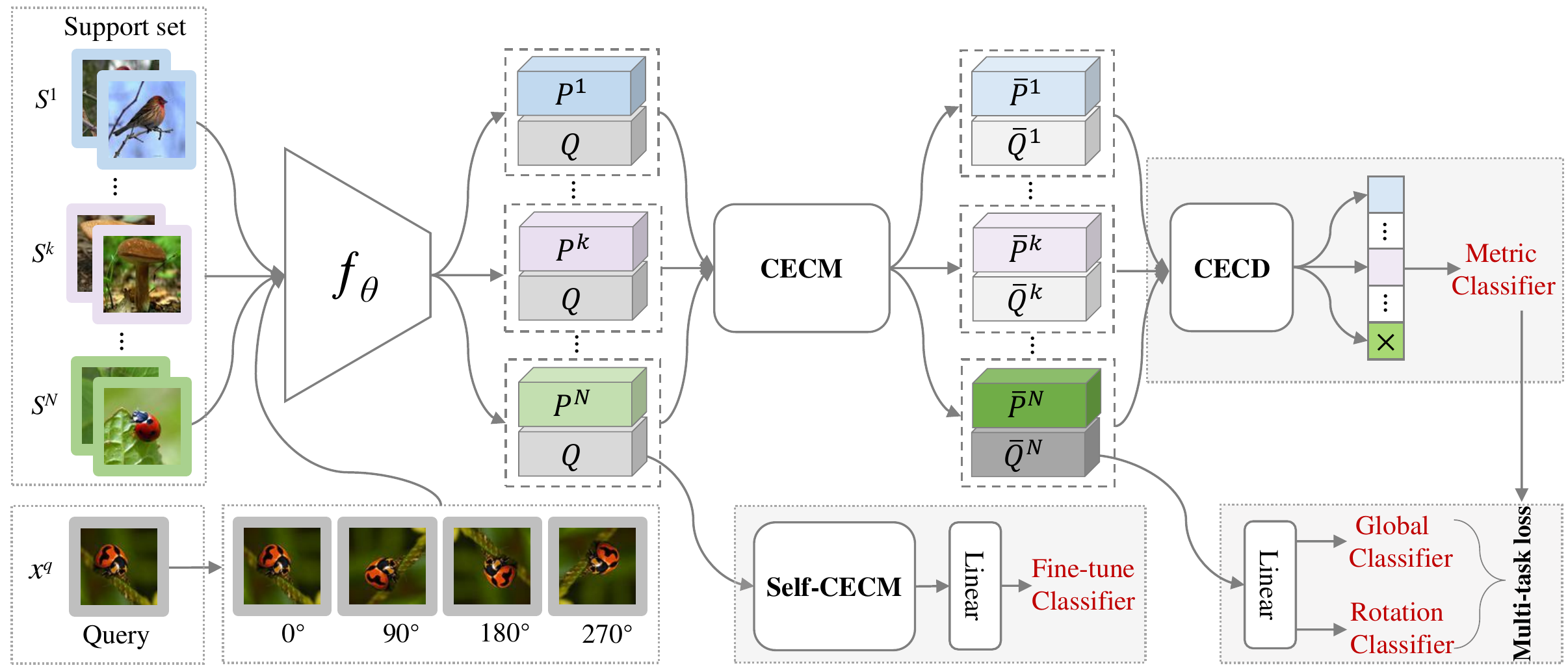}
%\vspace{-2mm}
\caption{The proposed CECNet framework. The CECM is able to highlight the mutually similar regions, the CECD is utilized to measure similarity of pair-features. And Self-CECM enhances the semantic feature of target object via self-connection.}
\label{fig:CECN}
%\vspace{-3mm}
\end{figure*}

\section{CEC Network for Few-Shot Classification}
%\vspace{-0.1cm}
\subsection{CEC Module and Self-CEC Module}
\label{subsec:cecm}
\seasons{According to the CEC layer mentioned above, we propose two derivative modules: the CEC Module (CECM) and the Self-CEC Module (self-CECM).}
The CECM is able to highlight the mutual similar regions via learning the semantic relevance between pair feature.
Specifically, CECM transfers the input pair-features ($P$, $Q$) $\in \mathbb{R}^{hw \times c}$ into more discriminative representations ($\bar{P}$, $\bar{Q}$) $\in \mathbb{R}^{hw \times c}$.
Formally, its function $f_{CECM}$ is expressed as:
\begin{equation}
\begin{aligned}
({\bar{Q}},{\bar{P}}) &= f_{CECM}(Q,P),\\
where, \quad \bar{Q}&=f_{CEC}(Q,P),\quad \bar{P}=f_{CEC}(P,Q).
\label{eq:cecm}
\end{aligned}
\end{equation}

The Self-CECM enhances the semantic feature of target object via self-connection, which turns the input $Q$ into $\bar{\bar{Q}} \in \mathbb{R}^{hw \times c}$.
\ljx{
Formally, Self-CECM function $f_{SCECM}$ is expressed as:
\begin{equation}
\begin{aligned}
{\bar{\bar{Q}}}=f_{SCECM}(Q)=f_{CEC}(Q,Q).
\label{eq:scecm}
\end{aligned}
\end{equation}
}\ljx{The CECM exploit the relation between P and Q via $\bar{Q}=f_{CEC}(Q,P)$, while Self-CECM exploit the relation between the input itself via $\bar{\bar{Q}}=f_{CEC}(Q,Q)$, i.e. Self-CECM explores the relation between the patches of input image. Because we assume that patch-features of the target are mutually similar, Self-CECM can enhance the target region by clustering the similar regions.}

\subsection{CECNet Framework}
Then, we give the overall Clustered-patch Element Connection Network (CECNet). The framework is shown in Fig.\ref{fig:CECN}, which integrates CECM, Metric Classifier and Fine-tune Classifier for few-shot classification task, and Rotation Classifier and Global Classifier for the auxiliary tasks.
\seasons{The network involves three stages: Base Training, Novel Fine-tuning and Novel Inference.}

\noindent\ljx{\textbf{Base Training}} \
As illustrated in Fig.\ref{fig:CECN}, every image $x^q$ in query set $\mathcal{Q}=\{\left(x^q_i, y^q_i\right)\}_{i=1}^{n_q}$ is rotated with [0\degree, 90\degree, 180\degree, 270\degree] and outputs a rotated $\mathcal{\tilde{Q}}=\{\left(\tilde{x}^q_i, \tilde{y}^q_i\right)\}_{i=1}^{{n_q}\times4}$.
The support subset $\mathcal{S}^k$ and the rotated query instance $\tilde{x}^q$ are processed by the embedding ${f_\theta}$ and produces the prototype feature $P^k=\frac{1}{|\mathcal{S}^k|} \sum_{x^s_i\in \mathcal{S}^k} {f_\theta}(x^s_i)$ and query feature ${Q}={f_\theta}(\tilde{x}^q)\in \mathbb{R}^{c\times h\times w}$, respectively. Then each pair-features ($P^k$, $Q$) are processed via CECM to enhance the mutually similar regions and generates more discriminative features ($\bar{P}^k$, $\bar{Q}^k$) for the subsequent classification. Note that the inputs and outputs of CECM will be reshaped to satisfied its format.
\jinxiang{Finally, CECNet is optimized via multi-task loss contributing from metric classifier and auxiliary tasks.}

\noindent\ljx{\textbf{Novel Fine-tuning}} \
The Fine-tune Classifier consists of Self-CECM and a linear layer as shown in Fig.\ref{fig:CECN}.
In fine-tuning phase, the pre-trained embedding ${f_\theta}$ is frozen, and the Fine-tune Classifier is optimized with cross-entropy loss.

\noindent\ljx{\textbf{Novel Inference}} \
In inductive inference, the overall prediction of CECNet is ${Y}={Y_M}+{Y_F}$, where ${Y_M}$ and ${Y_F}$ are the results of Metric and Fine-tune Classifiers respectively.

\subsection{Metric Classifier}
As illustrated in Eq.~\ref{eq:ec_Q}, the proposed CEC layer is able to generate a reliable relation map ${R^Q}$. The relation map ${R^Q}$ can also be utilized as a similarity map, and the mean of ${R^Q}$ is the similarity score.
Therefore, we obtain the CECD distance metric $d_{CECD}$ which is expressed as:
\begin{small}
\begin{equation}
\begin{aligned}
d_{CECD}(\bar{Q},\bar{P}) = \left(\frac{\bar{Q}}{||\bar{Q}||_2}\otimes \frac{C^{\bar{p}}}{||C^{\bar{p}}||_2}\right)\in \mathbb{R}^{hw}.
\label{eq:cecd}
\end{aligned}
\end{equation}
\end{small}

%As a metric-based meta learner, the Metric Classifier predicts the query into $N$ support categories by measuring similarity with the proposed CECD distance metric. Following \cite{hou2019cross}, we adopt the patch-wise classification mechanism to generate precise embeddings. Specifically, each local feature $\bar{Q}^k_n$ at $n^{th}$ spatial position of $\bar{Q}^k$, is predicted into $N$ categories. FoThe Metric Classifier make predictions by rmally, the probability of recognizing $\bar{Q}^k_n$ as $k^{th}$ category is:
\jinxiang{With the proposed CECD distance metric, the Metric Classifier make predictions by measuring the similarity between the query and the $N$ support classes.
Following \cite{hou2019cross}, the patch-wise classification strategy is used to produce precise feature representations.
In detail, each patch-wise feature $\bar{Q}^k_n$ at $n^{th}$ spatial position of $\bar{Q}^k$, is recognized as $N$ classes.
And the probability of predicting $\bar{Q}^k_n$ as $k^{th}$ class is:}
\begin{small}
\begin{equation}
\begin{aligned}
\hat{Y}(y=k|\bar{Q}^k_n)=\frac{\exp{(R^{k}_{n})}}
{\sum_{i=1}^{N} \exp{\left(R^{i}_{n}\right)}}, \\
where, \quad {R^{k}}=d_{CECD}(\bar{Q}^k,\bar{P}^k)\in \mathbb{R}^{hw},
\label{equ:pred}
\end{aligned}
\end{equation}
\end{small}where, the similarity map $R^{k}$ is obtained by the CECD distance metric formulated in Eq.~\ref{eq:cecd}, and the similarity score $R^{k}_{n}$ is the $n^{th}$ position of $R^{k}$.

\subsection{Fine-tune Classifier}
The Fine-tune Classifier consists of Self-CECM and a linear layer.
\seasons{It} predicts the query feature $\bar{\bar{Q}}$ into $N$ categories by a linear layer $W_F$.
And its loss is computed as:
\begin{small}
\begin{equation}
\begin{aligned}
\mathcal{L}_F&=PCE\left(W_F(\bar{\bar{Q}}),N^q\right) \\
&=-\sum_{i=1}^{n_q} \sum_{n=1}^{h \times w} {N^q_i} \log \left(\sigma\left(W_F(\bar{\bar{Q}}_n)_i\right)\right),
\end{aligned}
\label{equ:LF}
\end{equation}
\end{small}where, ${PCE}$ is patch-wise cross-entropy, and ${N^q_i}$ is the ground truth of ${x}^q_i$ with ${N}$ categories of few-shot task.

\renewcommand{\tabcolsep}{5.0pt}
\begin{table*}[t]
\centering
%\vspace{-0.2cm}
\begin{tabular}{ l | c | c c | c c}
\hline
\multicolumn{1}{l|}{\multirow{2}*{Model}}  & \multirow{2}*{Backbone} & \multicolumn{2}{c|}{miniImageNet}  &\multicolumn{2}{c}{tieredImageNet} \\
\cline{3-6}
\multicolumn{1}{c|}{ } & & 1-shot &5-shot &1-shot &5-shot \\
\hline
%MatchingNet \cite{vinyals2016matching} &Conv4 &43.44 $\pm$ 0.77 & 60.60 $\pm$ 0.71 & - & -\\
%MAML \cite{finn2017model} &Conv4 &48.70 $\pm$ 0.84 & 55.31 $\pm$ 0.73 & 51.67 $\pm$ 1.81 & 70.30 $\pm$ 1.75  \\
%MetaNet \cite{munkhdalai2017meta} &Conv4 & 49.21 $\pm$ 0.96 & - & - &- \\
ProtoNet \cite{snell2017prototypical} &Conv4 &49.42 $\pm$ 0.78 & 68.20 $\pm$ 0.66 &53.31 $\pm$ 0.89 &72.69 $\pm$ 0.74\\
%RelationNet \cite{sung2018learning} &Conv4 &50.44  $\pm$ 0.82 & 65.32  $\pm$ 0.70 & 54.48  $\pm$ 0.93 & 71.32  $\pm$ 0.78\\
%MM-Net \cite{qi2018memory} &Conv4 &53.37 $\pm$ 0.48 &66.97 $\pm$ 0.35 & - & -\\
\hdashline
\textbf{Our CECNet}  &Conv4 & \textbf{54.45 $\pm$ 0.47} &\textbf{70.57 $\pm$ 0.38} &\textbf{56.59 $\pm$ 0.50} &\textbf{72.86 $\pm$ 0.42} \\
\hline
%adaNet \cite{munkhdalai2018rapid} &ResNet-12 &56.88 $\pm$ 0.62 & 71.94 $\pm$ 0.57 & - &-\\
%TADAM \cite{oreshkin2018tadam} &ResNet-12 &58.50 $\pm$ 0.30 & 76.70 $\pm$ 0.30 &- & -\\
%MTL \cite{sun2019meta} &ResNet-12 &61.20 $\pm$ 1.80 & 75.50 $\pm$ 0.80 &- &- \\
%MetaOpt \cite{lee2019meta} &ResNet-12 &62.64 $\pm$ 0.62 &78.63 $\pm$ 0.46 & 65.99 $\pm$ 0.72 &81.56 $\pm$ 0.53\\
CAN \cite{hou2019cross} &ResNet-12 &63.85 $\pm$ 0.48 & 79.44 $\pm$ 0.34 &69.89 $\pm$ 0.51 &84.23 $\pm$ 0.37 \\
%P-Transfer \cite{zhiqiang2021partial} &ResNet-12 &64.21 $\pm$ 0.77 & 80.38 $\pm$ 0.59 &- &- \\
%MetaOpt+ArL \cite{hongguang2021rethink} &ResNet-12 &65.21 $\pm$ 0.58 &80.41 $\pm$ 0.49 &- &-\\
DeepEMD \cite{zhang2020deepemd} &ResNet-12 &65.91 $\pm$ 0.82 & 82.41 $\pm$ 0.56 &71.16 $\pm$ 0.87 &86.03 $\pm$ 0.58 \\
IENet \cite{rizve2021exploring} &ResNet-12 &66.82 $\pm$ 0.80 & 84.35 $\pm$ 0.51 &71.87 $\pm$ 0.89 &86.82 $\pm$ 0.58 \\
%infoPatch \cite{liu2021learning} &ResNet-12 &67.67 $\pm$ 0.45 & 82.44 $\pm$ 0.31 &71.51 $\pm$ 0.52 &85.44 $\pm$ 0.35 \\
%RFS \cite{tian2020rethinking} &SEResNet-12 &67.73 $\pm$ 0.63 & 83.35 $\pm$ 0.41 &72.55 $\pm$ 0.69 &86.72 $\pm$ 0.49 \\
DANet \cite{xu2021learning} &ResNet-12 &67.76 $\pm$ 0.46 & 82.71 $\pm$ 0.31 &71.89 $\pm$ 0.52 &85.96 $\pm$ 0.35 \\
MCL \cite{liu2022learning} &ResNet-12 &67.36 $\pm$ 0.20 & 83.63 $\pm$ 0.20  &71.76 $\pm$ 0.20 &86.01 $\pm$ 0.20 \\
Meta-DeepBDC \cite{jiangtao2022joint} &ResNet-12 &67.34 $\pm$ 0.43 & 84.46 $\pm$ 0.28 &72.34 $\pm$ 0.49 &\textbf{87.31 $\pm$ 0.32} \\
%COSOC* \cite{xu2021Rectifying} &ResNet-12 &69.28 $\pm$ 0.49 & \textbf{85.16 $\pm$ 0.42} &\textbf{73.57 $\pm$ 0.43} &\textbf{87.57 $\pm$ 0.10} \\
\hdashline
\textbf{Our CECNet}  &ResNet-12 &\textbf{69.32 $\pm$ 0.46} &\textbf{84.65 $\pm$ 0.32} &\textbf{73.14 $\pm$ 0.50} &86.88 $\pm$ 0.36 \\
\hline
%wDAE-GNN \cite{gidaris2019generating} &WRN-28 &61.07 $\pm$ 0.15 &76.75 $\pm$ 0.11 &68.18 $\pm$ 0.16 & 83.09 $\pm$ 0.12 \\
%LEO \cite{rusu2019meta} &WRN-28 &61.76 $\pm$ 0.08 &77.59 $\pm$ 0.12 & 66.33 $\pm$ 0.05 & 81.44 $\pm$ 0.09 \\
%wDAE \cite{gidaris2019generating} &WRN-28 &62.96 $\pm$ 0.15 & 78.85 $\pm$ 0.10 & 68.18 $\pm$ 0.16 & 83.09 $\pm$ 0.12 \\
PSST \cite{zhengyu2021pareto} &WRN-28 &64.16 $\pm$ 0.44 & 80.64 $\pm$ 0.32 &- &- \\
%SimpleShot \cite{yan2019simpleshot} &DenseNet-121 &64.29 $\pm$ 0.20 & 81.50 $\pm$ 0.14 &71.32 $\pm$ 0.22 &86.66 $\pm$ 0.15 \\
%FEAT \cite{ye2020few} &WRN-28 &65.10 $\pm$ 0.20 & 81.11 $\pm$ 0.14 &70.41 $\pm$ 0.23 &84.38 $\pm$ 0.16 \\
%CA \cite{afrasiyabi2019associative} &WRN-28 &65.92 $\pm$ 0.60 & 82.85 $\pm$ 0.55 &\textbf{74.40 $\pm$ 0.68} &86.61 $\pm$ 0.59 \\
%CAN \cite{hou2019cross} &WRN-28 &66.12 $\pm$ 0.47 & 80.43 $\pm$ 0.33 &71.04 $\pm$ 0.53 &84.92 $\pm$ 0.37 \\
DANet \cite{xu2021learning} &WRN-28 &67.84 $\pm$ 0.46 & 82.74 $\pm$ 0.31 &72.18 $\pm$ 0.52 &86.26 $\pm$ 0.35 \\
\hdashline
%\textbf{Our CECNet-metric} &WRN-28 & 69.89 $\pm$ 0.46 &84.20 $\pm$ 0.30 &73.01 $\pm$ 0.50 &86.62 $\pm$ 0.34 \\
\textbf{Our CECNet}  &WRN-28 & \textbf{70.20 $\pm$ 0.46}& \textbf{85.00 $\pm$ 0.30} & \textbf{73.84 $\pm$ 0.50} &\textbf{87.36 $\pm$ 0.34} \\
\hline
\end{tabular}
%\vspace{-0.4cm}
\caption{Comparing to existing approaches on 5-way FSL classification task on miniImageNet and tieredImageNet.
Our CECNet adopts the proposed CECM(T) attention module, CECD(C) distance metric, and Self-CECM.}
\label{table:SOTA}
\end{table*}

\subsection{Objective functions in Base Training}
\paragraph{Metric Loss}
The metric classification loss with the ground-truth few-shot label $\tilde{y}^q$ is:
\begin{small}
\begin{equation}
\mathcal{L}_M = -\sum_{i=1}^{n_q} \sum_{n=1}^{h \times w}\log \hat{Y}(y=\tilde{y}^q_i|(\bar{Q}_n)_i).
\label{equ:LM}
\end{equation}
\end{small}

\paragraph{Auxiliary Loss}
\jinxiang{The loss of Global Classifier is $\mathcal{L}_G=PCE(W_G(\bar{Q}),D^q)$, where ${D^q_i}$ is the global category of $\tilde{x}^q_i$ with $D$ classes of train set, and $W_G$ is a fully-connected layer.
Similarly, the loss of Rotation Classifier is ${\mathcal{L}_R=PCE(W_R(\bar{Q}),B^q)}$, where ${B^q_i}$ is the rotation category of $\tilde{x}^q_i$ with four classes, and $W_R$ is a fully-connected layer.}
%The Global Classifier predicts the query into all $D$ categories of train set, thus its loss is $\mathcal{L}_G=PCE(W_G(\bar{Q}),D^q)$, where $W_G$ is a linear layer, ${D^q_i}$ is the global category of $\tilde{x}^q_i$ with all $D$ categories.
%Similarly, the loss of Rotation Classifier is derived by ${\mathcal{L}_R=PCE(W_R(\bar{Q}),B^q)}$, where $W_R$ is a linear layer, and ${B^q_i}$ is the rotation category of $\tilde{x}^q_i$ with four categories.

\paragraph{Multi-Task Loss}
\ljx{Then, inspired by \cite{lai2022rethinking}, the overall loss is defined as:
\begin{small}
\begin{equation}
%%\vspace{-0.5mm}
\begin{aligned}
\mathcal{L} = \frac{1}{2}{\mathcal{L}_M} + \sum_{j=G,R}\left({\left({\lambda}+{w_j}\right)}{\mathcal{L}_j}+{log{\frac{1}{{({\lambda}+{w_j})}}}}\right),
\end{aligned}
\label{equ:Loss}
%%\vspace{-0.5mm}
\end{equation}
\end{small}where ${w} = \frac{1}{2{\alpha^2}}$ and ${\alpha}$ is a learnable variable.
The hyper-parameter ${\lambda}$ is utilized to balance the few-shot and auxiliary tasks, of which the influence is studied in Tab.~\ref{table:ablation_mtl}.
}

%\vspace{-0.1cm}
\section{Experiments on Few-Shot Classification}
%\vspace{-0.1cm}
\paragraph{Datasets}
\jinxiang{The two popular FSL classification benchmark datasets are \textit{mini}ImageNet and \textit{tiered}ImageNet, where detailed introductions are presented in APPENDIX.}

\paragraph{Experimental Setup}
We report the mean accuracy by testing $2000$ episodes randomly sampled from meta-test set.
According to Tab.~\ref{table:ablation_mtl}, the hyperparameter $\lambda$ is set to $1.0$ and $2.0$ for ResNet-12 and WRN-28, respectively.
Other implementation details can be found in our public code.

%\vspace{-0.1cm}
\subsection{Comparison with State-of-the-arts}
As shown in Tab.\ref{table:SOTA}, we compare with the state-of-the-art few-shot methods on miniImageNet and tieredImageNet datasets.
\jinxiang{It shows that our CECNet outperforms the existing SOTAs, which demonstrates the effectiveness and strength of our CEC based methods.}
Different from existing metric-based methods \cite{zhang2020deepemd,liu2022learning,jiangtao2022joint} extracting support and query features independently, our CECNet enhances the semantic feature regions of mutually similar objects and obtains more discriminative representations. Comparing to the metric-based Meta-DeepBDC \cite{jiangtao2022joint}, CECNet achieves $1.98\%$ higher accuracy on 1-shot.
Some metric-based methods \cite{xu2021learning,hou2019cross} apply cross attention, while our CECNet still surpasses DANet \cite{xu2021learning} with an accuracy improvement up to $2.36\%$ under WRN-28 backbone, which demonstrates the strength of our Clustered-patch Element Connection.

\renewcommand{\tabcolsep}{3.0pt}
\begin{table}[t]
\centering
%\vspace{-0.2cm}
\begin{tabular}{ c | a | c | c | c  c}
\hline
\multirow{2}*{Model}  & \multirow{1}*{Attention} & \multirow{1}*{Distance}& \multirow{2}*{Param}& \multicolumn{2}{c}{miniImageNet} \\
\cline{5-6}
 &Module &Metric & & 1-shot &5-shot \\
\hline
ProtoG &- &\multirow{2}*{cosine}&7.75M &61.87 &78.87  \\
CAN & CAM &&7.75M &63.85 &79.44  \\
\hline
\multirow{4}*{CECNet} &CECM(M) &\multirow{4}*{cosine}&7.75M &67.69 &81.84 \\
 & CECM(C) &&7.75M &67.65 & 81.79  \\
 & CECM(G) &&8.00M &{67.80} &{82.15}  \\
 & CECM(T) & &10.25M &\textbf{67.91} &\textbf{82.40}  \\
\hline
\end{tabular}
%\vspace{-0.4cm}
\caption{The 5-way classification results studying the influence of CECM with ResNet-12. In line with the setting of CAN, cosine distance metric is applied, and Rotation and Fine-tune classifications are not applied.
The CECM(M/C/G/T) denote different modes of Patch Cluster such as MatMul, Cosine, GCN and Transformer. Based on ProtoNet, ProtoG adds auxiliary global classification task.}
\label{table:ablation_cecm}
\end{table}

\renewcommand{\tabcolsep}{3.0pt}
\begin{table}[t]
\centering
%\vspace{-0.2cm}
\begin{tabular}{ c | c | a | c | c  c}
\hline
\multirow{2}*{Model}  & \multirow{1}*{Attention} & \multirow{1}*{Distance}& \multirow{2}*{Param}& \multicolumn{2}{c}{miniImageNet} \\
\cline{5-6}
&Module &Metric & & 1-shot &5-shot \\
\hline
ProtoG &- &cosine&7.75M &61.87 &78.87  \\
\hdashline
\multirow{4}*{CECNet} &\multirow{4}*{-} &CECD(M)&7.75M &67.50 &82.00  \\
&  &CECD(C)&7.75M &\textbf{67.89} &\textbf{82.02}  \\
&  &CECD(G)&8.00M &67.79 &81.74  \\
&  &CECD(T)&10.25M &67.44 &81.17  \\
\hline
\multirow{4}*{CECNet} &\multirow{4}*{CECM(T)} &CECD(M)&10.25M &67.64 &81.24 \\
&  &CECD(C)&10.25M &\textbf{68.27} & \textbf{82.59}  \\
&  &CECD(G)&11.25M &{66.52} &{78.55}  \\
&  &CECD(T) &12.75M &{64.37} &{78.32}  \\
\hline
\end{tabular}
%\vspace{-0.4cm}
\caption{The 5-way classification results studying the influence of CECD with ResNet-12. The setting is consistent with Tab.\ref{table:ablation_cecm}, except for distance metric.
The CECD(M/C/G/T) denote different modes such as MatMul, Cosine, GCN and Transformer.}
\label{table:ablation_cecd}
\end{table}

\renewcommand{\tabcolsep}{2.3pt}
\begin{table}[t]
\centering
%\vspace{-0.2cm}
\begin{tabular}{c | c c c | c  c | c  c}
\hline
\multirow{2}*{${\lambda}$} & \multicolumn{3}{c|}{Loss weights} & \multicolumn{2}{c|}{ResNet-12} & \multicolumn{2}{c}{WRN-28} \\
\cline{2-8}
&Metric &Global &Rotation & 1-shot &5-shot & 1-shot &5-shot \\
\hline
-&0.5 &- &- &62.45 &79.50 &61.98 &76.64 \\
-&0.5 &- &1.0 &65.54 &79.55 &63.47 &77.62 \\
-&0.5 &1.0 &- &68.27 &82.59 &67.13 &81.95 \\
-&0.5 &1.0 &1.0 &\textbf{68.86} &\textbf{83.67} &\textbf{69.49} &\textbf{83.71} \\
\hline
0.5&0.5 &${w_G}$ &${w_R}$ &{69.05} &{83.86} &69.33 &83.55 \\
1.0&0.5 &${w_G}$ &${w_R}$ &\textbf{69.32} &\textbf{84.21} &69.66 &84.09 \\
1.5&0.5 &${w_G}$ &${w_R}$ &69.15 &84.03 &69.86 &84.30  \\
2.0&0.5 &${w_G}$ &${w_R}$ &69.18 &83.29 &\textbf{70.20} &\textbf{84.59} \\
\hline
\end{tabular}
%\vspace{-0.6cm}
\caption{The 5-way classification results on \emph{mini}ImageNet studying the influence of multi-task loss applied in CECNet.}
\label{table:ablation_mtl}
\end{table}

\subsection{Ablation Study}
\paragraph{Influence of CECM}
As shown in Tab.\ref{table:ablation_cecm}, comparing CECNet to ProtoG, it shows consistent improvements on 1/5-shot classifications, because our CECM enhances the mutually similar regions and produces more discriminative representations. Comparing with CAN adopting cross attention module CAM, our CECNet achieves obvious improvements up to $4.06\%$ on 1-shot task. The results of CECM(M), CECM(C), CECM(G) and CECM(T) show that CECM is not sensitive to alternative modes such as MatMul, Cosine, GCN and Transformer, which indicates the generic Patch Cluster behavior is the key insight for the improvements.
%Even so, the Transformer version of Patch Cluster is still slightly outperform than others.

\paragraph{Influence of CECD}
As shown in Tab.\ref{table:ablation_cecd} without attention module, comparing CECNet to ProtoG, it shows consistent improvements, because our CECD distance metric can obtain a more reliable similarity map. Besides, the results show that the best combination is CECM(T) + CECD(C).

\paragraph{Influence of Multi-Task Loss}
In Tab.\ref{table:ablation_mtl} with the integration of auxiliary tasks, our CECNet obtains large improvements, which indicates that learning a good embedding is helpful.

\paragraph{Influence of CECM+CECD}
As shown in Tab.\ref{table:ablation_module}, comparing to ProtoG (no-attention + cosine), our methods adopting CECM(T) + cosine and no-attention + CECD(C) achieve obvious improvements, which demonstrates the effectiveness of the proposed CECM and CECD.
The combination of CECM(T) + CECD(C) obtains further performance gains.

\begin{figure}[!t]
\centering
\includegraphics[width=0.95\linewidth]{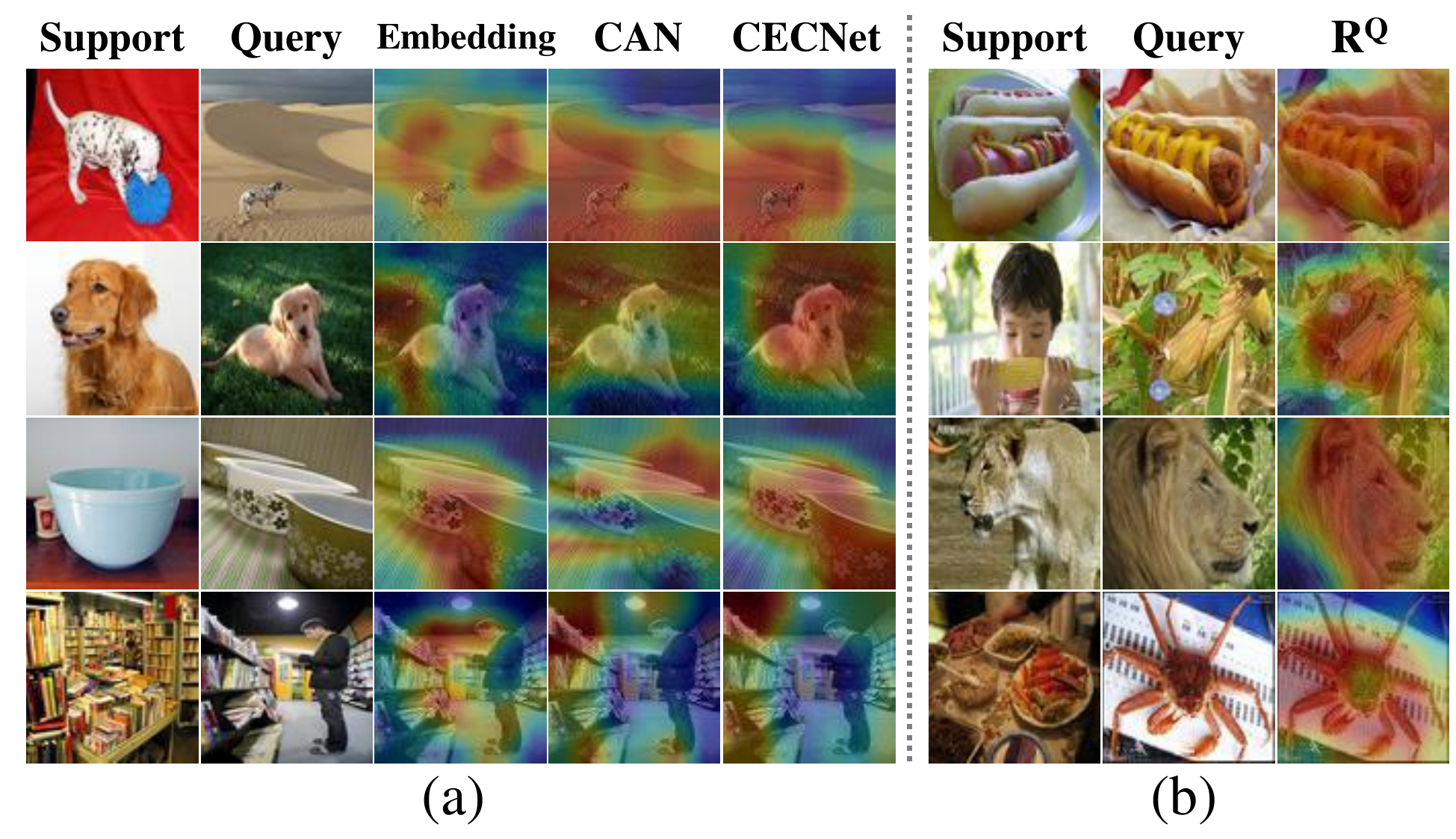}
%\vspace{-4mm}
\caption{(a) The class activation maps on 5-way 1-shot classification, where \textit{Embedding} belongs to CECNet.
(b) The visualizations of our CEC-based relation map $R^Q$.
}
\label{fig:visual}
%\vspace{-0.4cm}
\end{figure}

\renewcommand{\tabcolsep}{2pt}
\begin{table}[t]
\centering
%\vspace{-0.2cm}
\begin{tabular}{ c | c | c | c  c}
\hline
\multirow{1}*{Attention} & \multirow{1}*{Distance}& \multirow{2}*{Param}& \multicolumn{2}{c}{miniImageNet} \\
\cline{4-5}
Module &Metric & & 1-shot &5-shot \\
\hline
- &cosine &7.75M &65.59 $\pm$ 0.47 &80.94 $\pm$ 0.33  \\
CECM(T) &cosine &10.25M &68.27 $\pm$ 0.46 &83.43 $\pm$ 0.32  \\
- &CECD(C) &7.75M &68.79 $\pm$ 0.46 &83.39 $\pm$ 0.32  \\
CECM(T) &CECD(C) &10.25M &\textbf{69.32 $\pm$ 0.46} &\textbf{84.21 $\pm$ 0.32}  \\
\hline
\end{tabular}
%\vspace{-0.4cm}
\caption{The 5-way results studying the influence of CECM+CECD, under ResNet-12 applying multi-task loss with ${\lambda}=1.0$.}
\label{table:ablation_module}
\end{table}

\renewcommand{\tabcolsep}{2.0pt}
\begin{table}[t]
\centering
%\vspace{-0.2cm}
\begin{tabular}{c | c  c | c  c}
\hline
\multirow{1}*{Metric}  & \multicolumn{2}{c|}{Fine-tune Classifier} & \multicolumn{2}{c}{miniImageNet} \\
\cline{2-5}
classifier & Self-CECM& Linear& 1-shot &5-shot \\
\hline
\checkmark &- &- &\textbf{70.20 $\pm$ 0.46} &84.59 $\pm$ 0.30  \\
- &- &\checkmark &69.20 $\pm$ 0.47 &84.40 $\pm$ 0.30  \\
- &\checkmark &\checkmark &69.36 $\pm$ 0.46 &84.78 $\pm$ 0.30  \\
\checkmark &\checkmark &\checkmark &\textbf{70.20 $\pm$ 0.46} &\textbf{85.00 $\pm$ 0.30}  \\
\hline
\end{tabular}
%\vspace{-0.4cm}
\caption{The 5-way results of CECNet studying the influence of Self-CECM, under WRN-28 applying multi-task loss with ${\lambda}=2.0$.}
\label{table:ablation_self}
\end{table}

\renewcommand{\tabcolsep}{2.0pt}
\begin{table}[t]
\centering
%\vspace{-3mm}
\resizebox{0.95\columnwidth}{!}{
\begin{tabular}{l | cc | cc}
\hline
\multirow{2}{*}{Model} & \multicolumn{2}{c|}{PASCAL-$5^i$} & \multicolumn{2}{c}{COCO-$20^i$} \\
\cline{2-5}
&  1-shot & 5-shot & 1-shot & 5-shot \\
\hline
PPNet \cite{ppnet} & 51.5 & 62.0 & 25.7 & 36.2\\
RePRI \cite{Malik2021repri} & 59.3 & 64.8 & 36.6 & 45.2\\
%HSNet \cite{min2021hypercorrelation} & {64.0} & {69.5} & 39.2 & {46.9}\\
%CyCTR \cite{zhang2021few}           & {64.2} & 65.6 & 40.3 & 45.6\\
%DGPNet \cite{johnander2022dense}   & 63.5 & {73.5} & {45.0} & {56.2}\\
\hline
\textbf{RePRI+CECE(M)}     & 60.4 & \textbf{66.5} & \textbf{38.3} & \textbf{46.9}\\
\textbf{RePRI+CECE(T)}     & \textbf{60.5} & {66.2} & {38.1} & {46.7}\\
\hline
\end{tabular}
}
\caption{Comparison on PASCAL-$5^i$ and COCO-$20^i$ few-shot semantic segmentation benchmarks using mIoU with ResNet-50.
The CECE(M/T) denote different modes of MatMul and Transformer.}
\label{tab:FSSS_results}
%\vspace{-3mm}
\end{table}

\renewcommand{\tabcolsep}{2.0pt}
\begin{table}[!t]
\centering%
%\vspace{-3mm}
\resizebox{0.95\columnwidth}{!}{
\begin{tabular}{l | cc | cc}
\hline
\multirow{2}{*}{Model} & \multicolumn{2}{c|}{PASCAL} & \multicolumn{2}{c}{COCO} \\
\cline{2-5}
&  1-shot & 5-shot & 1-shot & 5-shot \\
\hline
DeFRCN \cite{qiao2021defrcn} & 52.5 & 60.7 & 6.5 & 15.3\\
MFDC \cite{wu2022multi} & 56.1 & 62.2 & 10.8 & 16.4\\
\hline
\textbf{MFDC+CECE(M)}     &\textbf{59.4}  & {63.4} & \textbf{11.5} & \textbf{17.2}\\
\textbf{MFDC+CECE(T)}     &58.7  & \textbf{64.9} & {11.2} & {16.9}\\
\hline
\end{tabular}
}
\caption{Comparison on PASCAL Novel Split 3 (nAP50) and COCO (nmAP) few-shot object detection benchmarks with ResNet-101.}
\label{tab:FSOD_results}
%\vspace{-5mm}
\end{table}

\paragraph{Influence of Self-CECM} As illustrated in Tab.\ref{table:ablation_self}, the baseline is the Metric Classifier of CECNet, and the competitor is Fine-tune Classifier with only Linear layer. By comparing Self-CECM+Linear to Linear, it shows consistent improvements, which demonstrates the usefulness of Self-CECM. By comparing Metric+Fine-tune to Metric Classifier, it shows an improvement on 5-shot classification.

%\vspace{-0.2cm}
\subsection{Visualization Analysis}
%\vspace{-0.1cm}
Fig.\ref{fig:visual}(a) shows the class activation maps \cite{zhou2016learning} of our CECNet and CAN \cite{hou2019cross}. Comparing CECNet to its \textit{Embedding}, CECNet can highlight the target object which is unseen in the pre-training stage.
Comparing to CAN, CECNet is more accurate and has larger receptive fields. The essential is that our Clustered-patch Element Connection utilizes the global info to implement element connection leading to a more confident correlation and a more clear connection.
Fig.\ref{fig:visual}(b) shows the visualizations of the CEC-based relation map $R^Q$ generated by CECNet via Eq.\ref{eq:ec_Q}. Our CEC approach produces a high-quality relation map with a more complete region for the target.

%\vspace{-0.2cm}
\section{Applications on FSSS and FSOD Tasks}
%\vspace{-0.1cm}
\jinxiang{In this section, we first introduce a novel CEC-based embedding module named CEC Embedding (CECE). Then, we extend the proposed CECE into few-shot semantic segmentation (FSSS) and object detection (FSOD) tasks. The experimental results in Tab.\ref{tab:FSSS_results} and Tab.\ref{tab:FSOD_results} show that our CECE can achieve performance improvements around $1\%-3\%$, and more extensive results are presented in APPENDIX.}

\paragraph{CEC Embedding}
%\jinxiang{Formally, the CEC Embedding $f_{CECE}$ is expressed as:
\jinxiang{$f_{CECE}$ is expressed as:
\begin{equation}
\begin{aligned}
{{Q}'} = f_{CECE}(Q) = f_{CEC}(Q,W_E).
\label{eq:cece}
\end{aligned}
\end{equation}
where, $\{Q, Q'\}\in \mathbb{R}^{hw \times c}$ are the input and output features respectively, and $W_E\in \mathbb{R}^{n_e \times c}$ are learnable weights (pytorch code is $W_E=nn.Embedding(n_e, c)$, and $n_e$ represents the number of semantic groups, and the empirical setting is $n_e=5$).
The proposed CECE can enhance the target regions of input features that are semantically similar to $W_E$, where $W_E$ contains the semantic information of base categories after trained on the base dataset.}

\paragraph{CECE Applications}
\jinxiang{As an embedding module, our CECE can be stacked after the backbone network.
To verify the effectiveness of the proposed CECE, we insert it into the FSSS method RePRI \cite{Malik2021repri} and FSOD method MFDC \cite{wu2022multi}, via stacking CECE after their backbones.
As illustrated in Tab.\ref{tab:FSSS_results} and Tab.\ref{tab:FSOD_results}, our CECE can make consistent improvements upon RePRI and MFDC methods.}

%\vspace{-0.2cm}
\section{Conclusion}
%\vspace{-0.1cm}
We propose a novel Clustered-patch Element Connection network (CECNet) for few-shot classification. Firstly, we design a Clustered-patch Element Connection (CEC) layer, which strengthens the target regions of query features by element-wisely connecting them with the clustered-patch features.
Then three useful CEC-based modules are derived: CECM and Self-CECM generate more discriminative features, and CECD distance metric obtains a reliable similarity map.
Extensive experiments prove that our method is effective, and achieves the state-of-the-arts on few-shot classification benchmark.
\jinxiang{Furthermore, our CEC approach can be extended into few-shot segmentation and detection tasks, which achieves competitive improvements.}

\appendix

\section{Related Work}
\paragraph{Few-Shot Classification}
The representative inductive few-shot classification approaches include parameter-generating based, optimization-based, metric-learning based, and embedding-based methods.
\textit{Parameter-generating methods} \cite{munkhdalai2017meta,gidaris2019generating} consider the model as a parameter generating network. \textit{Optimization-based methods} can rapidly adapt to unseen categories with a few samples by learning a well-initialized model \cite{nichol2018first,finn2017model} or a good optimizer \cite{marcin2018learn,ravi2016optimization}. \textit{Metric-learning based methods} \cite{vinyals2016matching,snell2017prototypical,xu2021learning,hou2019cross} classify an image by measuring similarity between it and the labeled samples. \textit{Embedding-based methods} \cite{tian2020rethinking,rizve2021exploring,zhengyu2021pareto} aim to learn a generalize-well embedding, then further fine-tune a linear classifier on novel categories.
%In a word, the propose few-shot classification methods lack of an uniform function to control the connections among the patches between support and query instances.

\paragraph{Auxiliary Task for FSL}
Recent works have demonstrated the effectiveness of introducing auxiliary tasks to FSL, which leads to a performance improvement via sharing parameters across tasks. In CAN \cite{hou2019cross}, a Global Classifier was proposed as the supervised auxiliary task for FSL. Some self-supervised based auxiliary tasks are applied for FSL, including contrastive learning \cite{liu2021learning}, rotation prediction \cite{su2020does} and geometric prediction \cite{rizve2021exploring}.

\paragraph{Few-Shot Semantic Segmentation}
Early few-shot semantic segmentation methods apply a dual-branch architecture \cite{shaban2017one,dong2018few,rakelly2018conditional}, one segmenting query-images with the prototypes learned by the other branch. In recently, the dual-branch architecture is unified into a single-branch, using the same embedding for support and query images \cite{zhang2020sg,siam2019amp,wang2019panet,rpmm,ppnet}. These methods aim to leverage better guidance for the segmentation of query-images \cite{zhang2020sg,nguyen2019feature,wangfew,zhang2019pyramid}, via learning better class-specific representations \cite{wang2019panet,liu2020crnet,ppnet,rpmm,siam2019amp} or iteratively refining \cite{zhang2019canet}.

\paragraph{Few-Shot Object Detection}
Existing few-shot object detection approaches can be divided into two paradigms: meta-learning based \cite{kang2019few,xiao2020few,fan2020few,hu2021dense} and transfer learning based \cite{wang2020frustratingly,wu2020multi,sun2021fsce,fan2021generalized,qiao2021defrcn,wu2022multi}.
The majority of meta-learning approaches adopt \emph{feature reweighting} or its variants to aggregate query and support features, which predict detections conditioned on support sets. Differently, the transfer learning based approaches firstly train the detectors on base set, then fine-tune the task-head layer on novel set, which achieve competitive results comparing to meta-learning approaches.

\paragraph{Graph Convolutional Network (GCN)}
GCN \cite{bruna2013spectral,atwood2016diffusion} extends convolution to graph domain.
Its representative directions are spectral-based \cite{defferrard2016convolutional,levie2018cayleynets} and spatial-based \cite{niepert2016learning,gilmer2017neural} methods.
In \cite{zonghan2020gcn}, more comprehensive reviews of GCN are introduced.

\paragraph{Transformer}
Transformer is an attention-based architecture with powerful learning ability, which has been applied in many computer vision tasks, such as classification \cite{wang2021pyramid}, detection\cite{zhu2020deformable} and segmentation\cite{zheng2021rethinking,liang2020polytransform}.

\renewcommand{\tabcolsep}{2.0pt}
\begin{table}[t]
\centering
\vspace{-0.2cm}
\begin{tabular}{ l | c | c c}
\hline
\multicolumn{1}{l|}{\multirow{2}*{Model}}  & \multicolumn{2}{c}{CIFAR-FS} \\
\cline{2-3}
\multicolumn{1}{c|}{ } & 1-shot &5-shot \\
\hline
RFS~\cite{tian2020rethinking} &71.50 $\pm$ 0.80 &86.00 $\pm$ 0.50 \\
MetaOpt~\cite{lee2019meta} &72.60 $\pm$ 0.70 &84.30 $\pm$ 0.50 \\
DSN-MR~\cite{simon2020adaptive} &75.60 $\pm$ 0.90 &86.20 $\pm$ 0.60 \\
IENet~\cite{rizve2021exploring} &76.83 $\pm$ 0.82 &89.26 $\pm$ 0.58 \\
\hdashline
\textbf{Our CECNet} &\textbf{79.67 $\pm$ 0.47} &\textbf{89.77 $\pm$ 0.32} \\
\hline
\end{tabular}
\vspace{-0.2cm}
\caption{Comparison on 5-way FSL classification on CIFAR-FS with ResNet-12 backbone.}
\label{table:SOTA_cifar}
\end{table}

\renewcommand{\tabcolsep}{6.0pt}
\begin{table}[ht]
%\vspace{-0.1cm}
\centering
%\vspace{-0.1cm}
\begin{tabular}{l | c | c  c}
\hline
\multicolumn{1}{c|}{\multirow{2}*{Model}}  & \multirow{2}*{Classifier}& \multicolumn{2}{c}{miniImageNet} \\
\cline{3-4}
 & & 1-shot &5-shot \\
\hline
COSOC \cite{xu2021Rectifying} &cosine &69.28 & 85.16  \\
COSOC+CECC &CECC &\textbf{69.76} &\textbf{85.63}  \\
\hline
\end{tabular}
%\vspace{-0.1cm}
\caption{The results on 5-way classification about the influence of CECC, with ResNet-12 backbone.}
\label{table:ablation_cecc}
\end{table}

\section{Comparison to Relevant Works}
(I) The FRN \cite{wertheimer2021few} and CTX \cite{doersch2020crosstransformers} are embedding modules, which uses support features to reconstruct or represent query feature. DeepEMD \cite{zhang2020deepemd} is a distance metric, which measures the similarity of pairs via the optimal matching cost. However, our CEC is a more general layer, which can be applied as an embedding module CECE, an attention module CECM or a distance metric CECD.
(II) Similar to traditional Cross Attention, DeepEMD performs Locala-to-Local fully connection, which still suffers the semantically inconsistent problem due to different scale objects. Differently, the essence of our CEC is performing the Global-to-Local element connection between the Clustered-patch $C_p$ (global) and query $Q$ (local). Besides, our Patch Cluster is a generic concept. Despite the introduced four solutions, both FRN and CTX may also be modified to perform Patch Cluster.
Specifically, CTX is similar to our MatMul mode, while FRN attempts to reconstruct $Q$ from $P$ via solving the linear least-squares problem $min(|Q-WP|)$.
(III) PaCa (Patch-to-Cluster Attention) \cite{grainger2022learning} performs self-clustering among patches of $P$, while our Patch Cluster performs reference-cluster among $P$ with $Q$ as the specific cluster center to  find semantic relationship.

\renewcommand{\tabcolsep}{5.0pt}
\begin{table*}[t]
\centering
\small
%\resizebox{\textwidth}{!}
{
\begin{tabular}{lcccccacccca}
    \toprule
     & & \multicolumn{5}{c}{1 shot} & \multicolumn{5}{c}{5 shot} \\
     \cmidrule(lr){3-7}\cmidrule(lr){8-12}
     Model & Backbone & Fold-0 & Fold-1 & Fold-2 & Fold-3 & Mean & Fold-0 & Fold-1 & Fold-2 & Fold-3 & Mean \\
     \midrule
     CANet \cite{zhang2019canet} & \multirow{5}{*}{ResNet-50} & 52.5 & 65.9 & 51.3 & \textbf{51.9} &  55.4 & 55.5 & 67.8 & 51.9 & 53.2  & 57.1 \\
     PGNet \cite{zhang2019pyramid}  &  &  56.0 & 66.9 & 50.6 & 50.4 &  56.0 & 57.7 & 68.7 & 52.9 & 54.6  & 58.5 \\
     %%CRNet \cite{liu2020crnet} &  & - & - & - & - & 55.7 & - & - & - & - & 58.8  \\
     %%SimProp \cite{gairola2020simpropnet} &  & 54.9 & 67.3 & 54.5 & {52.0} & 57.2& 57.2 & 68.5 & 58.4 & 56.1 & 60.0 \\
     %LTM \cite{yang2020new} & & 52.8 & \textbf{69.6} & 53.2 & \textbf{52.3} & 57.0& 57.9 & 69.9 & 56.9 & 57.5 & 60.6 \\

     RPMM \cite{rpmm} & &55.2 & 66.9 & 52.6 & 50.7 & 56.3 & 56.3 & 67.3 &  54.5 & 51.0 & 57.3  \\
     PPNet \cite{ppnet} & &47.8 & 58.8 & 53.8 & 45.6 & 51.5 & 58.4 & 67.8 &  64.9 &56.7 &  62.0  \\
     %PFENet \cite{pfenet} & & \textbf{61.7} & \textbf{69.5} & 55.4 & \textbf{56.3} & \textbf{60.8} & 63.1 & 70.7 & 55.8 & 57.9 & 61.9 \\
     {RePRI} \cite{Malik2021repri} & &  60.8 & 67.8 & 60.9 & 47.5 & 59.3 & 66.0 & 70.9 & 65.9 & 56.4 & 64.8 \\
     \hline
     \textbf{RePRI+CECE(M)} & \multirow{2}{*}{ResNet-50} &  \textbf{61.6} & {68.4} & {61.4} & {50.2} & {60.4} & {66.0} & \textbf{71.3} & \textbf{68.3} & \textbf{60.2} & \textbf{66.5} \\
     \textbf{RePRI+CECE(T)} & &  {61.5} & \textbf{68.7} & \textbf{62.2} & {49.5} & \textbf{60.5} & \textbf{66.7} & {70.9} & {68.1} & {59.1} & {66.2} \\
    \bottomrule\\
\end{tabular}
}
\caption{The results on 1-way PASCAL-5$^i$ few-shot semantic segmentation using mean-IoU.
The CECE(M) and CECE(T) denote different modes such as MatMul and Transformer adopted in Patch Cluster.}
\label{tab:FSSS_PASCAL_results}
%%\vspace{-2mm}
\end{table*}

\renewcommand{\tabcolsep}{5.0pt}
\begin{table*}[t]
\centering
\small
%\resizebox{\textwidth}{!}
{
\begin{tabular}{lcccccacccca}
\toprule
& & \multicolumn{5}{c}{1 shot} & \multicolumn{5}{c}{5 shot} \\
\cmidrule(lr){3-7}\cmidrule(lr){8-12}
Model & Backbone & Fold-0 & Fold-1 & Fold-2 & Fold-3 & Mean & Fold-0 & Fold-1 & Fold-2 & Fold-3 & Mean \\
\midrule
PPNet \cite{ppnet} & \multirow{4}{*}{ResNet-50} & 34.5 & 25.4 & 24.3 & 18.6 & 25.7 & \textbf{48.3} & 30.9 & 35.7 & 30.2 & 36.2\\
RPMM \cite{rpmm}  &  & 29.5 & 36.8 & 29.0 & 27.0 & 30.6 & 33.8 & 42.0 & 33.0 & 33.3 &  35.5\\
PFENet \cite{pfenet} &  &  {36.5} & {38.6} & {34.5} & {33.8} & {35.8} & 36.5 & 43.3 & 37.8 & 38.4 & 39.0 \\
{RePRI} \cite{Malik2021repri} & & 36.1 & 40.0 & 34.0 & 36.1 & 36.6 & {43.3} & {48.7} & {44.0} & {44.9} &{45.2} \\
\hline
\textbf{RePRI+CECE(M)} & \multirow{2}{*}{ResNet-50} &  \textbf{38.5} & {40.8} & \textbf{35.7} & \textbf{38} & \textbf{38.3} & {44.4} & {51.1} & \textbf{45.8} & \textbf{46.4} & \textbf{46.9} \\
 \textbf{RePRI+CECE(T)} &  &  {37.9} & \textbf{41.3} & {35.2} & {37.9} & {38.1} & {44.3} & \textbf{51.2} & {45.2} & {46.1} & {46.7} \\
\bottomrule
\end{tabular}
}
\caption{The results on 1-way COCO-20$^i$ few-shot semantic segmentation using mean-IoU.
The CECE(M) and CECE(T) denote different modes such as MatMul and Transformer adopted in Patch Cluster.}
\label{tab:FSSS_COCO_results}
\end{table*}

\renewcommand{\tabcolsep}{5.0pt}
\begin{table*}[t]
\centering
{
\resizebox{\textwidth}{!}
{
\begin{tabular}{l|l|l|c|ccccc|ccccc|ccccc}
\toprule[1.1pt]
\multicolumn{4}{l}{}  & \multicolumn{5}{c}{Novel Set 1}  & \multicolumn{5}{c}{Novel Set 2}  & \multicolumn{5}{c}{Novel Set 3}  \\
\cmidrule(lr){5-9}\cmidrule(lr){10-14}\cmidrule(lr){15-19}
\multicolumn{4}{l|}{\multirow{-2}{*}{Model}} & 1 & 2  & 3  & 5  & 10  & 1  & 2 & 3 & 5  & 10 & 1 & 2  & 3 & 5  & 10  \\

\midrule[0.9pt]
\multicolumn{4}{l|}{FRCN-ft \cite{yan2019meta}}       & 9.9         & 15.6          & 21.6          & 28.0          & 52.0          & 9.4           & 13.8          & 17.4          & 21.9          & 39.7          & 8.1           & 13.9          & 19.0          & 23.9          & 44.6          \\
\multicolumn{4}{l|}{FSRW \cite{kang2019few}}               & 14.2          & 23.6         & 29.8          & 36.5          & 35.6          & 12.3          & 19.6          & 25.1          & 31.4          & 29.8          & 12.5          & 21.3          & 26.8          & 33.8          & 31.0          \\

\multicolumn{4}{l|}{TFA \cite{wang2020frustratingly}}           & 25.3         & 36.4          & 42.1          & 47.9          & 52.8          & 18.3          & 27.          & 30.9          & 34.1          & 39.5          & 17.9          & 27.2          & 34.3          & 40.8          & 45.6          \\
\multicolumn{4}{l|}{FSDetView \cite{Xiao2020FSDetView}}          & 24.2          & 35.3          &  42.2         & 49.1          & 57.4       & 21.6         & 24.6          & 31.9       & 37.0         & 45.7         & 21.2         & 30.0         & 37.2         & 43.8         & 49.6  \\
\multicolumn{4}{l|}{{DeFRCN} \cite{qiao2021defrcn}}   & 40.2 & 53.6 & 58.2 & 63.6 & {66.5} & 29.5 & 39.7 & 43.4 & 48.1 & {52.8} & 35.0 & 38.3 & 52.9 & 57.7 & 60.8 \\
\multicolumn{4}{l|}{{MFDC} \cite{wu2022multi} }  & {63.4} & {64.7} & {66.3} & {69.4} & {68.1} & {41.8} & {45.5} & {51.9} & {53.8} & {51.7} & {56.1} & {58.3} & {59.0} & {62.2} & {63.7} \\
\midrule[0.9pt]
\rowcolor[HTML]{EFEFEF}
\multicolumn{4}{l|}{\textbf{MFDC+CECE(M)}}  & {64.1} & {65.7} & {66.8} & {69.5} & {68.4} & {42.2} & \textbf{47.6} & {53.3} & \textbf{55.0} & {53.1} & \textbf{59.4} & {60.9} & {60.4} & {63.4} & {65.3} \\
\rowcolor[HTML]{EFEFEF}
\multicolumn{4}{l|}{\textbf{MFDC+CECE(T)}}  & \textbf{64.4} & \textbf{66.4} & \textbf{67.9} & \textbf{71.2} & \textbf{69.3} & \textbf{42.3} & {46.7} & \textbf{53.9} & {54.6} & \textbf{53.6} & {58.7} & \textbf{63.3} & \textbf{60.7} & \textbf{64.9} & \textbf{66.4} \\
\bottomrule[1.1pt]
\end{tabular}}}
%%\vspace{-0.18cm}
\vspace{-0.2cm}
\caption{The few-shot object detection results on PASCAL VOC dataset. we evaluate the performance($nAP_{50}$) under ResNet-101 with \textit{G-FSOD} setting on three novel splits over multiple runs.
The CECE(M) and CECE(T) denote different modes such as MatMul and Transformer adopted in Patch Cluster.}
\label{tab:voc_result}
\end{table*}

\section{Few-Shot Classification Datasets}
%\paragraph{Datasets}
We conduct experiments on \textit{mini}ImageNet, \textit{tiered}ImageNet and CIFAR-FS datasets. Following \cite{hou2019cross}, the 100 categories of \textit{mini}ImageNet dataset are split into 64, 16 and 20 categories for train, validation and test respectively. The \textit{tiered}ImageNet dataset \cite{ren2018meta} consists of 608 categories, which are divided into 351, 97 and 160 categories for train, validation and test respectively.
CIFAR-FS dataset randomly splits 100 classes of CIFAR-100 into 64, 16, and 20 categories corresponding to train, validation, and test.

\section{Comparison on CIFAR-FS}
As shown in Tab.~\ref{table:SOTA_cifar}, we compare with the state-of-the-art few-shot methods on CIFAR-FS datasets.
\jinxiang{It shows that our CECNet outperforms the existing SOTAs, which demonstrates the effectiveness and strength of our CECNet.}

\section{CEC Classifier}
Despite the derived CEC-based four modules, such as CECM and Self-CECM attention modules, CECD distance metric, and CECE embedding module, we further introduce a novel CEC Classifier (CECC) for few-shot learning.

Our CECNet is mete-learning based approach, while the recent works show that supervised-learning based few-shot classification methods \cite{chen2019closer,tian2020rethinking,xu2021Rectifying} also achieve very competitive accuracy performance.
In \cite{chen2019closer}, a cosine classifier is introduced for supervised-learning based few-shot classification. Inspired by the cosine classifier \cite{chen2019closer}, our CECD(C) distance metric can be modified into a Clustered-patch Element Connection Classifier (CECC), formally:
\setcounter{equation}{16}
\begin{equation}
\begin{small}
\begin{aligned}
CECC(Q)=CECD(W,Q)= \left(\frac{W}{||W||_2}\otimes \frac{C^q}{||C^q||_2}\right)\in \mathbb{R}^{D},
\label{eq:cecc}
\end{aligned}
\end{small}
\end{equation}
where, ${W \in \mathbb{R}^{D \times c}}$ are learnable weights, ${D}$ is the all categories of train set.

\subsection{COSOC+CECC}
The COSOC \cite{xu2021Rectifying} approach achieves very competitive performance in few-shot classification, which adopts a complicated multi-stage framework, including pre-training backbone by contrastive learning, data clustering, generating cropped data, training backbone by cosine classifier and inference with Shared Object Concentrator (SOC).
Based on COSOC method, we replace the cosine classifier with the proposed CECC, which obtains the COSOC+CECC approach. The results in Tab. \ref{table:ablation_cecc} show that our CECC is able to boost the performance upon COSOC.

\renewcommand{\tabcolsep}{1.5pt}
\begin{table}[t]
\vspace{-0.1cm}
\centering
\vspace{-0.1cm}
\resizebox{0.475\textwidth}{!}
{\begin{tabular}{l|l|l|c|cccccc}
%\hline
\toprule[1.0pt]
\multicolumn{4}{l}{}  & \multicolumn{6}{c}{Shot Number} \\
\cmidrule(lr){5-10}
\multicolumn{4}{l|}{\multirow{-2}{*}{Model}} & 1        & 2              & 3             & 5             & 10    &30                \\ \midrule[0.9pt]
\multicolumn{4}{l|}{FRCN-ft \cite{yan2019meta}}          & 1.7         & 3.1  & 3.7          & 4.6          & 5.5          & 7.4                  \\
\multicolumn{4}{l|}{TFA \cite{wang2020frustratingly}}            & 1.9         & 3.9  & 5.1          & 7.0          & 9.1          & 12.1                 \\
\multicolumn{4}{l|}{FSDetView \cite{Xiao2020FSDetView}}           & 3.2         & 4.9  & 6.7         & 8.1         & 10.7        & 15.9           \\
\multicolumn{4}{l|}{{DeFRCN} \cite{qiao2021defrcn}}   & 4.8 & 8.5 & 10.7 & {13.6} & {16.8} & {21.2}  \\
\multicolumn{4}{l|}{{MFDC} \cite{wu2022multi} } & {10.8} & {13.9} & {15.0} & {16.4} & {19.4} & {22.7}  \\
\midrule[0.9pt]
\rowcolor[HTML]{EFEFEF}
\multicolumn{4}{l|}{\textbf{MFDC+CECE(M)}} & \textbf{11.5} & \textbf{14.6} & {15.4} & \textbf{17.2} & \textbf{19.7} & \textbf{22.9}  \\
\rowcolor[HTML]{EFEFEF}
\multicolumn{4}{l|}{\textbf{MFDC+CECE(T)}} & {11.2} & {14.2} & \textbf{15.6} & {16.9} & {19.6} & {22.8}  \\
\bottomrule[1.0pt]
\end{tabular}}
\vspace{-0.2cm}
\caption{The few-shot object detection results on COCO dataset. we report the performance ($nmAP$) under ResNet-101 with \textit{G-FSOD} setting over multiple runs.
The CECE(M) and CECE(T) denote different modes such as MatMul and Transformer adopted in Patch Cluster.}
\label{tab:coco_result}
\end{table}

\section{Applications on FSSS and FSOD Tasks}
As shown in Tab.\ref{tab:FSSS_PASCAL_results}, Tab.\ref{tab:FSSS_COCO_results}, Tab.\ref{tab:voc_result}, and Tab.\ref{tab:coco_result}, our CECE can achieve performance improvements around $1\%-3\%$ on few-shot semantic segmentation and object detection tasks.

\paragraph{FSSS: Datasets and Setting}
\emph{PASCAL-5$^i$ and COCO-20$^i$ Datasets:}
PASCAL-5$^i$ is built from PASCAL VOC \cite{everingham2010pascal}.
The 20 object categories are split into 4 folds. For each fold, 15 categories are utilized for training and the remaining 5 classes for testing.
COCO-20$^i$ is built from MS-COCO \cite{ms-COCO}. COCO-20$^i$ dataset is divided into 4 folds with 60 base classes and 20 test classes in each fold.

\emph{Evaluation Setting:}
Following \cite{ppnet}, the mean Intersection over Union (mIoU) is adopted for evaluation, and we report average mIoU over 5 runs of 1000 tasks.

\paragraph{FSOD: Datasets and Setting}
\emph{PASCAL VOC and COCO Datasets:}
PASCAL VOC \cite{everingham2010pascal} are randomly sampled into 3 splits, and each contains 20 categories. For each split, there are 15 base and 5 novel categories.
Each novel class has $K = 1,2,3,5,10$ objects sampled from the train/val set of VOC2007 and VOC2012 for training, and the test set of VOC2007 for testing.
COCO \cite{lin2014microsoft} use 60 categories disjoint with VOC as base set, and the remaining 20 categories are novel set with $K = 1,2,3,5,10,30$ shots. The total 5k images randomly sampled from the validation set are utilized for testing, while the rest for training.

\emph{Evaluation Setting:} Following \cite{wang2020frustratingly,qiao2021defrcn}, we conduct experiments on the evaluation setting of generalized few-shot object detection \textit{\textbf{(G-FSOD)}}.

\bibliographystyle{named}
\bibliography{ijcai23}

\end{document}